\documentclass[letterpaper, 10pt, conference]{ieeeconf}
\IEEEoverridecommandlockouts   
\usepackage{cite}
\usepackage{amsmath,amssymb,amsfonts}
\usepackage{graphicx}
\usepackage{textcomp}
\usepackage{xcolor}
\usepackage{booktabs}
\usepackage{comment}
\usepackage[caption=false,font=footnotesize]{subfig}
\usepackage{caption}
\usepackage{url}

\usepackage{soul}

\makeatletter
\let\oldthebibliography\thebibliography
\renewcommand{\thebibliography}[1]{\oldthebibliography{99}}
\makeatother

\newcommand{\pmstd}[1]{\,{\scriptsize$\pm$#1}}
\newcommand{\predfirst}{\texttt{predfirst+enclast}}
\newcommand{\predlast}{\texttt{predlast+enclast}}

\def\BibTeX{{\rm B\kern-.05em{\sc i\kern-.025em b}\kern-.08em
    T\kern-.1667em\lower.7ex\hbox{E}\kern-.125emX}}
\begin{document}
\bstctlcite{IEEEexample:BSTcontrol}

\title{\LARGE \bf Sandwich-Residuals: Parameter-Efficient Test-time Adaptation of World Models}


\author{%
Krishnam Soni,
Aditya Sehgal,
Vedant Dave$^{\dagger}$,
and Elmar Rueckert$^{\dagger}$%
\thanks{%
\hspace*{-1em}%
$^{\dagger}$Vedant Dave and Elmar Rueckert contributed equally to the supervision of this work.\newline
All authors are with the Chair of Cyber-Physical-Systems,\newline
Montanuniversität Leoben, Austria.\newline
Corresponding author: \texttt{krishnamsoni07@gmail.com}
}%
}

\maketitle
\thispagestyle{empty}
\pagestyle{empty}

\begin{abstract}
Latent world models enable planning by predicting the effects of actions in a learned representation space, but their predictions can become unreliable when test-time conditions differ from training. Existing test-time adaptation methods address this by updating parts of the pretrained model, often modifying millions of parameters and requiring a choice of which internal components to adapt. We introduce \emph{Sandwich-Residuals}, a lightweight alternative that keeps the pretrained world model frozen and learns only small residual corrections around the predictor. The residuals are optimized online using the model's self-supervised prediction error and require no rewards, labels, or source-domain data. Across 21 conditions on the AdaJEPA benchmark, our method achieves \(1.3\times\) the success rate of the frozen model while retaining 95\% of the performance of the strongest AdaJEPA variant and adapting 97--99\% fewer parameters. Under compound shifts, this advantage increases to \(1.9\times\) the success rate of the frozen model, while remaining comparable to internal block adaptation. We further demonstrate the same adaptation principle on a DINO-WM model for 3-D manipulation. These results suggest that effective test-time adaptation of world models does not necessarily require modifying their pretrained internal weights.
 Project Page: \url{https://sandwich-residuals.github.io/}
\end{abstract}

\section{INTRODUCTION}
\label{sec:intro}
World models provide a compact mechanism for learning how an environment evolves under an agent's actions. Rather than reasoning directly in high-dimensional observation space, latent world models encode observations into structured representations and predict their future evolution conditioned on actions~\cite{ha2018worldmodels, Hafner2020Dream, hansen2024td}. Joint-Embedding Predictive
Architectures (JEPAs)~\cite{assran2023self, sobal2022joint} extend this idea by learning action-conditioned latent dynamics from reward-free trajectories and have recently shown strong results
for visual model-predictive control~\cite{zhou2025dinowm,wang2026temporal,maes2026leworldmodel}.

A central assumption behind such models is that a frozen world model remains valid at test-time, i.e., that the environment encountered at test time stays sufficiently close to the distribution on which it was trained.  In realistic robotic settings, this assumption is easily violated, and train--test distribution shifts can substantially degrade learned predictions and downstream control~\cite{sinha2022system}. Such shifts may affect observations through changes in lighting, backgrounds, camera viewpoint, or task-irrelevant distractors, without changing the underlying task~\cite{wang2026adajepa, toso2026learning}. In addition, the physical interaction between actions and state transitions may also change: friction, mass, actuator response, payload, or other system properties can alter how the same action affects the environment~\cite{kumar2021rma}. These two forms of distribution shift pose fundamentally different challenges. Appearance shifts perturb the observation mapping while potentially preserving the dynamics, whereas dynamics shifts directly invalidate the learned action-conditioned transition model.

Existing approaches address distribution shift either by learning representations that generalize beyond the training distribution, e.g., through state abstraction, bisimulation, or disentangled world models, or by adapting pretrained models using data encountered at test-time~\cite{zhou2022domain,liang2025comprehensive}. For observation-side shifts, Toso et al.~\cite{toso2026learning} explicitly shape the latent geometry of a JEPA-based world model using bisimulation, suppressing visually irrelevant variation while preserving transition-relevant structure and thereby improving planning robustness to appearance changes.

Test-time adaptation provides a natural way to correct test-time mismatch using transitions collected online. AdaJEPA~\cite{wang2026adajepa}, for example, updates a predictor block and the visual encoder's projection head using the model's own latent prediction error. While effective, this strategy modifies millions of pretrained parameters and requires selecting which internal predictor block to adapt. Wang et al.~\cite{wang2026adajepa} find that the best adaptation target is environment-dependent, although performance is generally not highly sensitive to the particular layers adapted. Their LoRA variant, which inserts low-rank adapters into every linear layer of the predictor and encoder while keeping the pretrained weights frozen, also improves over the frozen model but does not consistently outperform direct updates to selected layers. In our experiments, the two selected-layer AdaJEPA variants likewise differ in effectiveness: \predfirst\ outperforms \predlast\ on aggregate, yet both require updating nearly \(10^7\) pretrained parameters. These results leave a more basic question unresolved: \emph{does effective test-time adaptation require modifying the pretrained world model itself, and if not, where can the necessary correction be introduced?}

We investigate whether test-time adaptation can be achieved without modifying the pretrained world model itself. As illustrated in Fig.~\ref{fig:architecture_comparison}, our approach freezes all encoder and predictor parameters and inserts lightweight residual modules around the predictor, forming a ``sandwich'' of learnable corrections before
and after the frozen dynamics model. On the input side, residuals modify the visual and action embeddings before they are passed to the predictor; the action residual is additionally conditioned on proprioception so that the same command can be corrected differently depending on the current system state. On the output side, residuals correct the predicted visual and proprioceptive latents before they are reused for rollout and planning. Only these residual modules are updated online, using the same self-supervised latent prediction error as AdaJEPA and requiring no rewards, labels, or source-domain data. We call this architecture \emph{Sandwich-Residuals}.

We evaluate Sandwich-Residuals on an expanded AdaJEPA-based evaluation suite using the released checkpoints and environments~\cite{wang2026adajepa}, comprising 21
primary and 7 compound conditions, and on a DINO-WM model trained for OGBench-Cube manipulation~\cite{park2025ogbench}. The evaluation covers changes in dynamics, appearance, object-shape, layout, and compound shifts. Our key contributions are:
\begin{itemize}
    \item We introduce \emph{Sandwich-Residuals}, a test-time adaptation method that keeps the pretrained encoder and predictor frozen and updates only lightweight residual modules.

    \item We show that interface-level adaptation recovers most of the
    performance gain of AdaJEPA while updating 97--99\% fewer parameters,  with particularly strong results under dynamics and compound shifts and without requiring predictor-block selection.

    \item We demonstrate that the same adaptation principle transfers from AdaJEPA-style world models to a DINO-WM model on 3-D robotic
    manipulation, using a single residual design across architectures.
\end{itemize}

\section{RELATED WORK}
\label{sec:related}

\subsection{Latent World Models and Predictive Planning}
World models learn compact predictive models of environment dynamics that
support planning or policy learning without repeatedly interacting with the
environment~\cite{ha2018worldmodels,Hafner2020Dream,hansen2024td}. Recent
work increasingly performs prediction directly in learned representation
spaces, avoiding expensive pixel reconstruction~\cite{assran2023self,
zhou2025dinowm,maes2026leworldmodel}. DINO-WM~\cite{zhou2025dinowm}
predicts future frozen DINOv2 features conditioned on actions and plans
directly in that latent space, whereas Wang et
al.~\cite{wang2026temporal} train the encoder and predictor jointly under a
curvature penalty that straightens latent trajectories to make them easier
to plan through. Our work concerns how such action-conditioned world models
should adapt when test-time conditions change.

\subsection{Test-Time Adaptation}
Test-time training and adaptation update pretrained models using unlabeled observations encountered after test-time~\cite{sun2020test,wang2021tent,
niu2022efficient}. Existing methods adapt models through self-supervised objectives, entropy minimization, selective parameter updates, or test-time feature and classifier adjustment~\cite{iwasawa2021ttclassifier,
niu2022efficient,niu2023towards}. While effective, online gradient-based adaptation can introduce substantial computational overhead and may degrade pretrained representations when large parts of the model are updated~\cite{niu2022efficient,niu2023towards}. This has motivated
increasingly lightweight adaptation strategies that restrict test-time
updates to selected parameters or auxiliary components~\cite{song2023ecotta}. Our work follows this direction in latent world models, but confines adaptation to lightweight residual modules at the predictor's input and output interfaces rather than modifying the pretrained encoder or predictor.

\subsection{Online Adaptation of World Models}
World-model adaptation has been studied through online fine-tuning, system
identification, and dynamics-specific updates~\cite{kumar2021rma,
levy2026simdist}. SimDist~\cite{levy2026simdist} freezes the transferred
representation, reward, and value models while fine-tuning the latent
dynamics from real-world transitions, explicitly preserving a stationary
latent target during adaptation. AdaJEPA~\cite{wang2026adajepa} brings
test-time adaptation directly into JEPA-based MPC: after executing an
action, the resulting transition provides a self-supervised prediction
target for updating selected encoder and predictor parameters before
replanning.

\subsection{Residual and Action-Specific Dynamics Adaptation}
Residual dynamics models provide a complementary strategy for correcting
model mismatch while preserving a pretrained model~\cite{kaufmann2023champion,
lanier2026redraw}. ReDRAW~\cite{lanier2026redraw} is closest to our
setting: it freezes a pretrained world model and learns a residual
correction to its latent-state dynamics from a small offline target-domain
dataset. Recent work has also emphasized explicitly structuring
action-conditioned dynamics. AdaWorld~\cite{gao2025adaworld} learns
transferable latent action representations, while
DWM~\cite{zhang2026dwm} separates action-driven transitions from
action-independent world effects during training. These approaches improve
transfer or dynamics modeling, but correct the model before or between
deployments rather than online inside the planning loop, and do not jointly
correct the action and observation pathways.

\section{METHODOLOGY}
\label{sec:sandwich_residuals}

\begin{figure*}[t]
\vspace*{5pt}
\centering
\includegraphics[width=\textwidth]{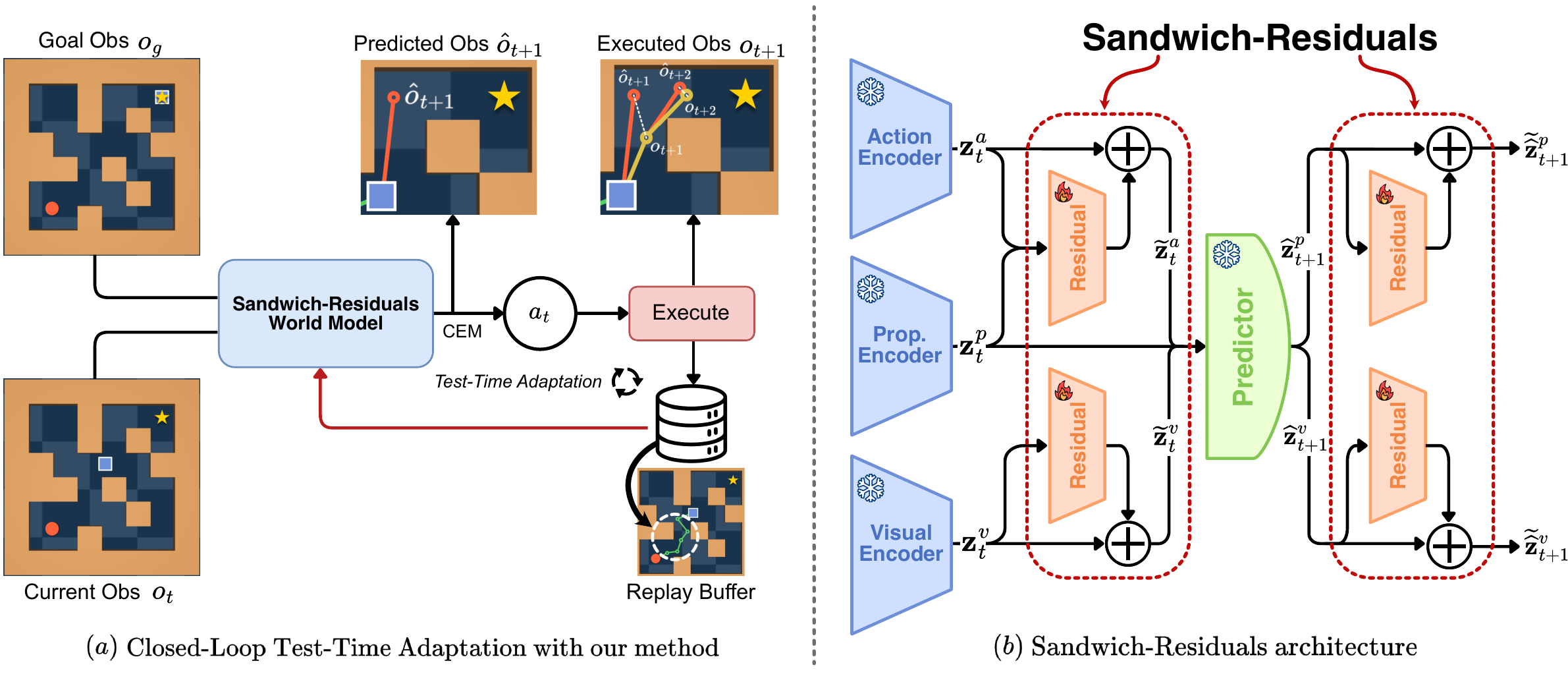}
\caption{\textbf{Overview and mechanism of Sandwich Residuals (SR).} \textbf{Left:} At each replanning step, the world model predicts the latent
representation of the next observation under the selected action \(a_t\),
shown schematically as \(\hat{o}_{t+1}\), while executing that action yields
the observed next observation \(o_{t+1}\). As test-time adaptation progresses,
updates from these observed transitions reduce the prediction mismatch,
illustrating progressive online recalibration of the model. \textbf{Right:} Circled \(+\) symbols denote additive residual connections. The orange residual blocks are adapted during test-time adaptation, while the encoders and predictor remain frozen. See Section~\ref{sec:sandwich_residuals} for notation and variable definitions.}
\label{fig:architecture_comparison}
\end{figure*}

\subsection{Background}
\label{sec:background}

We consider pretrained joint-embedding predictive architectures (JEPAs) for
visual model-predictive control. At time step \(t\), the agent receives an
observation \(\mathbf{o}_t\) and executes an action \(\mathbf{a}_t\).
Observation and action encoders produce latent representations that are
concatenated along their feature dimension into the predictor input
\(\mathbf{x}_t\). For a context of \(K\) consecutive time steps, the causal
predictor produces a one-step-ahead prediction at each context position:
\begin{equation}
\widehat{\mathbf{x}}_{t-K+2:t+1}
=
f_{\Theta}
\left(
\mathbf{x}_{t-K+1:t}
\right).
\end{equation}
During recursive rollout, the final prediction
\(\widehat{\mathbf{x}}_{t+1}\) is appended to the context before predicting
the next step.

We consider two instantiations of this architecture. In the pretrained
world models of AdaJEPA, the observation
\(\mathbf{o}_t=(\mathbf{v}_t,\mathbf{p}_t)\) consists of an image
\(\mathbf{v}_t\) and a proprioceptive state \(\mathbf{p}_t\), encoded by a
small ResNet with an MLP projection head and by a proprioceptive encoder,
respectively:
\begin{equation}
\mathbf{z}^{v}_t = \mathcal{E}_{v}(\mathbf{v}_t), \qquad
\mathbf{z}^{p}_t = \mathcal{E}_{p}(\mathbf{p}_t), \qquad
\mathbf{z}^{a}_t = \mathcal{E}_{a}(\mathbf{a}_t),
\end{equation}
with \(\mathbf{z}^{v}_t\in\mathbb{R}^{384}\) and
\(\mathbf{z}^{p}_t,\mathbf{z}^{a}_t\in\mathbb{R}^{10}\), and
\begin{equation}
\mathbf{x}_t
=
\left[
\mathbf{z}^{v}_t;\,
\mathbf{z}^{p}_t;\,
\mathbf{z}^{a}_t
\right].
\end{equation}
In DINO-WM, the observation is the image alone, encoded by a frozen DINOv2~\cite{oquab2024dinov2} backbone into a set of patch tokens
\(\mathbf{z}^{v}_t\in\mathbb{R}^{P\times384}\); the action embedding is
tiled onto every token, so that
\(\mathbf{x}_t=[\mathbf{z}^{v}_t;\,\mathbf{z}^{a}_t]\) per token, and the
predictor operates over the \(K\times P\) tokens of the context. In the
following, \(\mathbf{z}^{p}_t\) and the proprioceptive terms are simply
absent for DINO-WM.

The world model is trained using an objective centered on predicting future
observation representations in latent space. Its complete pretraining
objective also includes the auxiliary losses and regularization terms used
by the underlying JEPA formulation. After training, candidate action
sequences can be evaluated by recursively applying the predictor and
measuring the distance between the predicted representations and the
encoded goal observation. We use the cross-entropy method (CEM) to optimize
these action sequences within a receding-horizon control loop.

AdaJEPA adapts the pretrained world model during this control loop using
recently observed transitions. For clarity, consider the one-step
prediction ending at time \(i+1\). The observation components of the final
prediction produced by
\(f_{\Theta}(\mathbf{x}_{i-K+1:i})\) are concatenated as
\begin{equation}
\widehat{\mathbf{z}}^{o}_{i+1}
=
\left[
\widehat{\mathbf{z}}^{v}_{i+1};
\widehat{\mathbf{z}}^{p}_{i+1}
\right].
\label{eq:predicted_observation}
\end{equation}
The corresponding target representation is formed by encoding the observed
image and proprioceptive state:
\begin{equation}
\mathbf{z}^{o}_{i+1}
=
\left[
\mathcal{E}_{v}(\mathbf{v}_{i+1});
\mathcal{E}_{p}(\mathbf{p}_{i+1})
\right].
\label{eq:target_observation}
\end{equation}
Using the predicted and target observation representations in
Eqs.~\eqref{eq:predicted_observation}
and~\eqref{eq:target_observation}, respectively, AdaJEPA minimizes the
self-supervised latent prediction loss
\begin{equation}
\mathcal{L}_{\mathrm{pred}}
=
\frac{1}{|\mathcal{B}|}
\sum_{(\mathbf{o}_i,\mathbf{a}_i,\mathbf{o}_{i+1})\in\mathcal{B}}
\ell
\left(
\widehat{\mathbf{z}}^{o}_{i+1},
\operatorname{sg}\left(\mathbf{z}^{o}_{i+1}\right)
\right),
\label{eq:adaptation_loss}
\end{equation}
where \(\mathcal{B}\) contains the five most recent observed transitions,
\(\ell\) is the mean-squared error over the observation features, and
\(\operatorname{sg}(\cdot)\) denotes stop-gradient. The
\texttt{predfirst+enclast} variant updates the first predictor transformer
block, whereas \texttt{predlast+enclast} updates the last predictor
transformer block and the final predictor LayerNorm. Both variants also
update the encoder's final projection head where one exists; the frozen
DINOv2 encoder has none. Although this allows the world model to adapt to
test-time conditions, it requires modifying parameters of the pretrained
encoder and predictor.

\subsection{Residual Adaptation Before and After Prediction}
\label{sec:residual_adaptation}

As illustrated in Fig.~\ref{fig:architecture_comparison}, we freeze every parameter of the pretrained encoders and predictor and
introduce lightweight residual modules around the predictor. The residuals
modify its inputs and outputs, forming a sandwich around the frozen dynamics
model. This provides learnable interfaces for compensating for both
input-side distribution shifts and systematic prediction errors without
changing the pretrained world model itself.

We correct the action embedding using the action and, when available, the
proprioceptive representation:
\begin{equation}
\widetilde{\mathbf{z}}^{a}_t
=
\mathbf{z}^{a}_t
+
r^{a}_{\phi}
\left(
\left[
\mathbf{z}^{a}_t;
\mathbf{z}^{p}_t
\right]
\right),
\end{equation}
where \(r^{a}_{\phi}\) is a linear residual map. Conditioning the correction
on the proprioceptive representation allows the same encoded action to be
adjusted differently depending on the current state. This input-side
correction is intended to compensate for dynamics shifts, such as changes
in damping, that alter how a given action affects the next state.

We also apply a residual MLP to the visual representation:
\begin{equation}
\widetilde{\mathbf{z}}^{v}_t
=
\mathbf{z}^{v}_t
+
r^{v,\mathrm{in}}_{\phi}
\left(
\mathbf{z}^{v}_t
\right).
\end{equation}
This residual is applied immediately after the visual encoder, so it acts on
every encoded image: the context frames, the observed next frame that
serves as adaptation target, and the goal image used by the planning
objective. Predictions and goal are therefore compared in the same
corrected representation, similar to AdaJEPA. For DINO-WM the residual is applied to each
patch token. 

We do not modify the input proprioceptive embedding, which reflects our setting. The proprioceptive observation mapping remains unchanged, while distribution shifts affect the visual observations or the mapping from actions to state transitions. As in
Sec.~\ref{sec:background}, the equations below are written for the AdaJEPA models; for DINO-WM the \(\mathbf{z}^{p}_t\) terms and the proprioceptive
output residual are absent.

The corrected predictor input is therefore
\begin{equation}
\widetilde{\mathbf{x}}_t
=
\left[
\widetilde{\mathbf{z}}^{v}_t;\,
\mathbf{z}^{p}_t;\,
\widetilde{\mathbf{z}}^{a}_t
\right].
\end{equation}
For a context of \(K\) time steps, the frozen predictor produces
\begin{equation}
\widehat{\mathbf{x}}_{t-K+2:t+1}
=
f_{\Theta}
\left(
\widetilde{\mathbf{x}}_{t-K+1:t}
\right).
\end{equation}
Let \(\widehat{\mathbf{z}}^{v}_{t+1}\) and
\(\widehat{\mathbf{z}}^{p}_{t+1}\) denote the visual and proprioceptive
components of a predicted representation. A second pair of residual modules corrects these predictor outputs:
\begin{align}
\widetilde{\widehat{\mathbf{z}}}^{v}_{t+1}
&=
\widehat{\mathbf{z}}^{v}_{t+1}
+
r^{v,\mathrm{out}}_{\phi}
\left(
\widehat{\mathbf{z}}^{v}_{t+1}
\right), \\
\widetilde{\widehat{\mathbf{z}}}^{p}_{t+1}
&=
\widehat{\mathbf{z}}^{p}_{t+1}
+
r^{p,\mathrm{out}}_{\phi}
\left(
\widehat{\mathbf{z}}^{p}_{t+1}
\right).
\end{align}
The output visual residual is another MLP, while the output
proprioceptive residual is linear. These output residuals directly
compensate for errors that remain after prediction. The corrected
representations are used both recursively during model rollouts and by the
latent planning objective.

Both visual residuals are two-layer MLPs with a 64-dimensional hidden layer
and GELU activation,
\(\mathbb{R}^{384}\rightarrow\mathbb{R}^{64}\rightarrow\mathbb{R}^{384}\).
For the AdaJEPA models, the action-input residual maps
\(\mathbb{R}^{20}\rightarrow\mathbb{R}^{10}\), while the
proprioceptive-output residual maps
\(\mathbb{R}^{10}\rightarrow\mathbb{R}^{10}\), yielding 99{,}520 trainable
parameters in total. This corresponds to 1.0\% of the 9.93M parameters
updated by AdaJEPA in the maze models and 2.7\% of the 3.66M updated in the
pushing models. For DINO-WM, the action residual maps
\(\mathbb{R}^{10}\rightarrow\mathbb{R}^{10}\) and the proprioceptive-output
residual is omitted, resulting in 99{,}310 trainable parameters, or 3.1\% of
the 3.23M parameters updated by AdaJEPA.

For Sandwich-Residuals, Eq.~\eqref{eq:adaptation_loss} is evaluated using
the residual-corrected prediction
\(\widetilde{\widehat{\mathbf z}}^{o}_{i+1}
=[\widetilde{\widehat{\mathbf z}}^{v}_{i+1};
\widetilde{\widehat{\mathbf z}}^{p}_{i+1}]\)
and target
\(\widetilde{\mathbf z}^{o}_{i+1}
=[\widetilde{\mathbf z}^{v}_{i+1};\mathbf z^{p}_{i+1}]\),
with stop-gradient applied to the target.

The linear residuals are initialized to zero and the last layer of each
MLP residual to \(\mathcal{N}(0,10^{-4})\), so the adapted model coincides
with the frozen pretrained model at the beginning of each episode. Only
the residual parameters \(\phi\) are optimized at test-time. After each
plan-execute cycle, we perform one Adam update:
\begin{equation}
\phi
\leftarrow
\phi
-
\eta_{\phi}
\nabla_{\phi}
\mathcal{L}_{\mathrm{pred}}.
\end{equation}
In all experiments, we use a learning rate of \(10^{-2}\) for the linear
residuals and \(5\times10^{-4}\) for the visual MLPs. The adaptation buffer
retains the five most recent transitions, and the prediction loss is
averaged over all windows of up to \(K\) consecutive transitions formed
from them. Residual parameters persist across replanning cycles within an episode and
are reset to their initial values at the beginning of each new episode,
mirroring AdaJEPA's protocol.

\section{EXPERIMENTS}
\label{sec:experiments}
We ask whether adapting only the Sandwich-Residuals recovers planning
performance under test-time shifts as effectively as AdaJEPA's
updates to the pretrained weights, and how the answer depends on the
type of shift.

\subsection{AdaJEPA Benchmark}
\label{sec:exp_adajepa}
\noindent\textbf{Environments and pretrained models:}
We use the released AdaJEPA checkpoints, environments, and planner; the
only code change is the addition of the residual arm. The benchmark
comprises four goal-conditioned visual control tasks, each with its own
pretrained world model: \emph{Medium Maze}, a MuJoCo point-mass agent
navigating a fixed maze; \emph{Diverse Maze}, the same agent on held-out
layouts; \emph{PushT}, a pymunk pusher moving a T-shaped block to a goal
pose (model trained on the T block only); and \emph{PushObj}, the same
pusher with a model trained on four shapes \{T, L, Z, +\}. Episodes are
goal-conditioned on a goal image (and proprioception); start/goal pairs
are sampled as in AdaJEPA (maze: cells at least 3 apart, or shortest-path
distance 3--5 for held-out layouts; pushing: two states 25 steps apart
on held-out trajectories). Success is defined as in AdaJEPA.\newline\newline
\textbf{Test-time shifts:}
Shifts are applied at test time only and fall into three groups.
\emph{Dynamics shifts} change the simulator: joint damping $\times50$ and
body density $0.2$--$10\times$ in Medium Maze; the pusher's PD
controller gain $k_v$ doubled in PushT. \emph{Appearance shifts} alter
the rendered images (applied identically to observed and goal frames):
Gaussian blur ($\sigma=2$), salt-and-pepper noise (1\%), brightness
scaled by 0.9, and recoloring of the agent, block, or goal anchor to
red. \emph{Task shifts} change what must be manipulated or where:
PushObj evaluates on 3 unseen shapes (I, small T, square) in addition to
the 4 trained shapes, and Diverse Maze uses unseen layouts. Seven
compound conditions combine a corruption with a dynamics change. In
total we evaluate 21 main and 7 compound conditions.\newline\newline
\textbf{Planning protocol:}
All arms share AdaJEPA's receding-horizon loop. The world model observes
one image every 5 simulation steps (5 low-level actions per block) with
context $K=3$. Each cycle, CEM optimizes 5 action blocks (200
candidates, 30 elites, 10 iterations) against the distance between the
predicted latent trajectory and the encoded goal; the first block is
executed, the rest warm-start the next cycle, for up to 20 cycles.\newline\newline
\textbf{Baselines:}
\emph{Frozen} plans with the pretrained model and no adaptation. \emph{AdaJEPA} adapts a subset of pretrained weights after every executed block with one Adam step on $\mathcal{L}_{\mathrm{pred}}$.
\texttt{predlast+enclast} updates the last predictor block, its final LayerNorm, and the encoder's projection head; \texttt{predfirst+enclast}
updates the first predictor block and the projection head (learning rates as in the original work; 9.93M parameters in the maze models, 3.66M in the pushing models). \emph{Ours} freezes all pretrained weights
and updates only the residual modules of
Sec.~\ref{sec:residual_adaptation} (99{,}520 parameters), with the same buffer, update schedule, and per-episode reset. Every (condition, arm)
cell is evaluated on 50 episodes for each of 5 seeds; we additionally report paired differences and their standard errors in
addition to per-condition means.

\begin{table}[t]
\vspace*{5pt}
\centering
\footnotesize
\setlength{\tabcolsep}{3pt}
\caption{Success rate (\%), mean\,$\pm$\,standard deviation over 5 seeds of 50 episodes. \texttt{pl}: \texttt{predlast+enclast}; \texttt{pf}: \texttt{predfirst+enclast}; $^{*}$: shape unseen during training; $\Delta$: gain of ours over the frozen model in percentage. Bold indicates the best method per condition.}

\label{tab:main}
\begin{tabular}{lccccr}
\toprule
Condition & Frozen & \texttt{pl} & \texttt{pf} & Ours & $\Delta$ \\
\midrule
\multicolumn{6}{l}{\textbf{Medium Maze}} \\
No shift & 80.8\,{\scriptsize$\pm$8.9} & \textbf{86.0}\,{\scriptsize$\pm$5.8} & 84.0\,{\scriptsize$\pm$5.7} & 81.2\,{\scriptsize$\pm$6.7} & \textcolor{blue}{+0.4} \\
Damping 50$\times$ & 46.4\,{\scriptsize$\pm$3.8} & 52.0\,{\scriptsize$\pm$4.0} & 59.2\,{\scriptsize$\pm$3.0} & \textbf{66.0}\,{\scriptsize$\pm$3.7} & \textcolor{blue}{+19.6} \\
Density 0.2$\times$ & \textbf{90.0}\,{\scriptsize$\pm$4.5} & 86.4\,{\scriptsize$\pm$6.2} & 88.0\,{\scriptsize$\pm$7.7} & 88.4\,{\scriptsize$\pm$5.7} & \textcolor{blue}{$-$1.6} \\
Density 10$\times$ & 38.0\,{\scriptsize$\pm$5.1} & 39.6\,{\scriptsize$\pm$6.1} & \textbf{42.8}\,{\scriptsize$\pm$4.6} & 40.4\,{\scriptsize$\pm$3.0} & \textcolor{blue}{+2.4} \\
Blur $\sigma$=2 & 83.2\,{\scriptsize$\pm$6.6} & 83.2\,{\scriptsize$\pm$6.4} & \textbf{84.0}\,{\scriptsize$\pm$4.7} & 83.2\,{\scriptsize$\pm$7.7} & \textcolor{blue}{+0.0} \\
\quad\textit{average} & \textit{67.7} & \textit{69.4} & \textit{71.6} & \textit{71.8} & \textcolor{blue}{\textit{+4.2}} \\
\midrule
\multicolumn{6}{l}{\textbf{Diverse Maze}} \\
Unseen layouts & 45.6\,{\scriptsize$\pm$9.1} & 53.6\,{\scriptsize$\pm$6.1} & 62.8\,{\scriptsize$\pm$8.2} & \textbf{63.2}\,{\scriptsize$\pm$11.0} & \textcolor{blue}{+17.6} \\
\midrule
\multicolumn{6}{l}{\textbf{PushObj}} \\
T & 50.4\,{\scriptsize$\pm$4.3} & 76.4\,{\scriptsize$\pm$3.0} & \textbf{79.6}\,{\scriptsize$\pm$5.2} & 76.4\,{\scriptsize$\pm$6.5} & \textcolor{blue}{+26.0} \\
L & 47.6\,{\scriptsize$\pm$6.2} & 76.8\,{\scriptsize$\pm$3.0} & \textbf{77.2}\,{\scriptsize$\pm$2.3} & 71.6\,{\scriptsize$\pm$7.3} & \textcolor{blue}{+24.0} \\
Z & 37.6\,{\scriptsize$\pm$8.9} & 72.8\,{\scriptsize$\pm$4.6} & \textbf{79.2}\,{\scriptsize$\pm$6.7} & 70.4\,{\scriptsize$\pm$4.8} & \textcolor{blue}{+32.8} \\
Plus & 36.4\,{\scriptsize$\pm$5.5} & 72.0\,{\scriptsize$\pm$5.8} & \textbf{76.8}\,{\scriptsize$\pm$6.4} & 70.4\,{\scriptsize$\pm$8.2} & \textcolor{blue}{+34.0} \\
I$^{*}$ & 24.8\,{\scriptsize$\pm$7.0} & \textbf{46.0}\,{\scriptsize$\pm$9.4} & 43.2\,{\scriptsize$\pm$7.6} & 45.6\,{\scriptsize$\pm$8.9} & \textcolor{blue}{+20.8} \\
Small T$^{*}$ & 49.6\,{\scriptsize$\pm$4.6} & \textbf{65.6}\,{\scriptsize$\pm$9.0} & 63.2\,{\scriptsize$\pm$8.8} & 62.4\,{\scriptsize$\pm$11.8} & \textcolor{blue}{+12.8} \\
Square$^{*}$ & 22.4\,{\scriptsize$\pm$7.1} & 40.0\,{\scriptsize$\pm$5.1} & \textbf{44.4}\,{\scriptsize$\pm$7.0} & 39.2\,{\scriptsize$\pm$9.5} & \textcolor{blue}{+16.8} \\
\quad\textit{average} & \textit{38.4} & \textit{64.2} & \textit{66.2} & \textit{62.3} & \textcolor{blue}{\textit{+23.9}} \\
\midrule
\multicolumn{6}{l}{\textbf{PushT}} \\
No shift & 62.0\,{\scriptsize$\pm$6.0} & 75.6\,{\scriptsize$\pm$3.6} & \textbf{82.4}\,{\scriptsize$\pm$2.6} & 77.6\,{\scriptsize$\pm$5.5} & \textcolor{blue}{+15.6} \\
Blur $\sigma$=2 & 50.4\,{\scriptsize$\pm$4.3} & 69.2\,{\scriptsize$\pm$5.8} & \textbf{82.8}\,{\scriptsize$\pm$5.6} & 72.8\,{\scriptsize$\pm$4.1} & \textcolor{blue}{+22.4} \\
S\&P noise 0.01 & 55.6\,{\scriptsize$\pm$4.1} & 70.4\,{\scriptsize$\pm$8.8} & \textbf{83.2}\,{\scriptsize$\pm$2.3} & 72.0\,{\scriptsize$\pm$4.2} & \textcolor{blue}{+16.4} \\
Brightness 0.9$\times$ & 65.6\,{\scriptsize$\pm$7.1} & 74.8\,{\scriptsize$\pm$6.3} & \textbf{82.0}\,{\scriptsize$\pm$4.2} & 73.6\,{\scriptsize$\pm$6.5} & \textcolor{blue}{+8.0} \\
Red agent & 50.0\,{\scriptsize$\pm$4.7} & 62.4\,{\scriptsize$\pm$6.1} & \textbf{74.0}\,{\scriptsize$\pm$4.9} & 67.2\,{\scriptsize$\pm$3.3} & \textcolor{blue}{+17.2} \\
Red block & \textbf{14.0}\,{\scriptsize$\pm$4.2} & 11.2\,{\scriptsize$\pm$4.1} & 10.0\,{\scriptsize$\pm$2.0} & 12.8\,{\scriptsize$\pm$1.8} & \textcolor{blue}{$-$1.2} \\
Red anchor & 29.6\,{\scriptsize$\pm$3.8} & 36.0\,{\scriptsize$\pm$5.7} & \textbf{43.2}\,{\scriptsize$\pm$8.2} & 42.4\,{\scriptsize$\pm$3.3} & \textcolor{blue}{+12.8} \\
$k_v$ 2$\times$ & 53.2\,{\scriptsize$\pm$5.4} & 80.8\,{\scriptsize$\pm$5.2} & \textbf{91.2}\,{\scriptsize$\pm$3.0} & 88.4\,{\scriptsize$\pm$5.4} & \textcolor{blue}{+35.2} \\
\quad\textit{average} & \textit{47.5} & \textit{60.0} & \textit{68.6} & \textit{63.4} & \textcolor{blue}{\textit{+15.8}} \\
\midrule
\textbf{All 21 conditions} & 49.2 & 63.4 & \textbf{68.2} & 65.0 & \textcolor{blue}{\textbf{+15.8}} \\
\bottomrule
\end{tabular}
\end{table}
 
\noindent\textbf{Results:}
Tables~\ref{tab:main} and~\ref{tab:compound} summarize performance on the
21 primary and seven compound conditions, respectively, while
Table~\ref{tab:paired} reports the paired comparisons. Across the 21 primary conditions, our method
achieves 65.0\% mean success, compared with 49.2\% for Frozen, 63.4\%
for \texttt{predlast+enclast}, and 68.2\% for
\texttt{predfirst+enclast}. This is achieved while adapting only
1.0\% of the parameters updated by AdaJEPA in the maze models and
2.7\% in the pushing models. Overall, our method improves over Frozen by
$15.8\pm1.2$ and over \texttt{predlast} by $1.6\pm0.7$, while remaining
within $3.2\pm0.8$ of \texttt{predfirst}.

\begin{figure}[t]
    \vspace*{5pt}
    \centering
    \includegraphics[width=0.95\columnwidth]{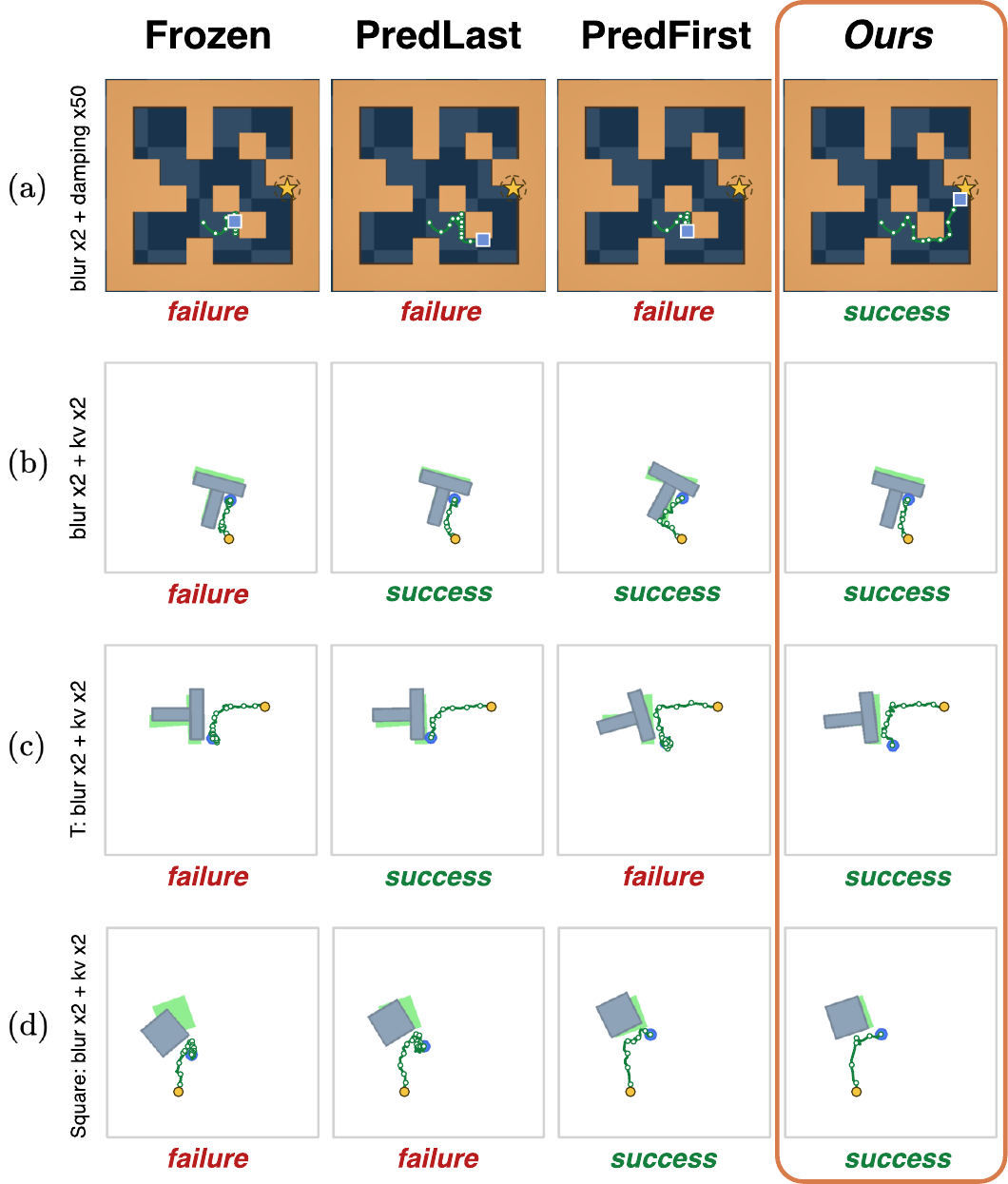}
    \caption{Example episodes under compound shifts, planned with each of
    the four models, from the same start state and goal:
    (a) Medium Maze, blur $\sigma=2$ + damping $50\times$; (b) PushT, blur
    $\sigma=2$ + controller $k_v$ $2\times$; (c) PushObj T, blur $\sigma=2$
    + controller $k_v$ $2\times$; (d) PushObj square, blur $\sigma=2$ +
    controller $k_v$ $2\times$. Dotted green lines trace the executed
    trajectory. Frozen fails in all four episodes; ours succeeds in all
    four, uniquely reaching the goal under $50\times$ damping in (a), while
    \texttt{predfirst} fails in (c) and \texttt{predlast} fails in (d).
    These are illustrative single episodes and not representative of the
    aggregate rates in Tables~\ref{tab:main} and~\ref{tab:compound}.}
    \label{fig:trajectories}
    \vspace{-0.5cm}
\end{figure}

\begin{table*}[t]
\vspace*{5pt}
\centering
\small
\setlength{\tabcolsep}{10pt}
\caption{Success rate (\%) under compound shifts, reported as
mean\,$\pm$\,standard deviation over 5 seeds of 50 episodes.
\texttt{pl} and \texttt{pf} are defined in Table~\ref{tab:main}.
Bold indicates the best method per condition.
$\Delta$ denotes the gain of ours over the frozen model in percentage.}
\label{tab:compound}
\begin{tabular}{@{}lccccr@{}}
\toprule
Condition
& Frozen
& \texttt{pl}
& \texttt{pf}
& Ours
& \textcolor{blue}{$\Delta$} \\
\midrule

Medium Maze: blur $\sigma=2$ + damping 50$\times$
& 46.8\pmstd{7.8}
& 54.0\pmstd{4.7}
& 56.0\pmstd{4.0}
& \textbf{65.6}\pmstd{6.2}
& \textcolor{blue}{+18.8} \\

Medium Maze: blur $\sigma=2$ + density 10$\times$
& 38.8\pmstd{6.1}
& 39.2\pmstd{3.6}
& 37.2\pmstd{7.2}
& \textbf{40.8}\pmstd{3.0}
& \textcolor{blue}{+2.0} \\

PushT: blur $\sigma=2$ + controller $k_v$ 2$\times$
& 40.4\pmstd{4.3}
& 76.8\pmstd{3.3}
& \textbf{87.6}\pmstd{2.6}
& \textbf{87.6}\pmstd{3.0}
& \textcolor{blue}{+47.2} \\

PushT: red anchor + controller $k_v$ 2$\times$
& 30.4\pmstd{4.3}
& 42.4\pmstd{7.4}
& \textbf{53.2}\pmstd{4.1}
& 51.2\pmstd{7.7}
& \textcolor{blue}{+20.8} \\

PushT: blur $\sigma=2$ + red anchor + controller $k_v$ 2$\times$
& 34.4\pmstd{8.6}
& 41.2\pmstd{5.8}
& \textbf{68.8}\pmstd{6.7}
& 59.6\pmstd{2.6}
& \textcolor{blue}{+25.2} \\

PushObj T: blur $\sigma=2$ + controller $k_v$ 2$\times$
& 26.8\pmstd{3.0}
& 84.8\pmstd{2.7}
& 85.6\pmstd{3.3}
& \textbf{87.2}\pmstd{5.0}
& \textcolor{blue}{+60.4} \\

PushObj Square: blur $\sigma=2$ + controller $k_v$ 2$\times$
& 23.6\pmstd{6.4}
& 66.4\pmstd{8.6}
& 68.0\pmstd{9.1}
& \textbf{70.0}\pmstd{9.1}
& \textcolor{blue}{+46.4} \\
\bottomrule
\end{tabular}
\end{table*}

\begin{table}[h]
\centering
\small
\caption{Paired differences in success rate (percentage points), computed
as ours minus each baseline and reported as mean\,$\pm$\,standard error over pairs. \texttt{pl} and \texttt{pf} are defined in Table~\ref{tab:main}.}
\label{tab:paired}
\begin{tabular*}{\columnwidth}{
  @{\extracolsep{\fill}}
  lrrr
  @{}
}
\toprule
Subset & $-$ Frozen & $-$ \texttt{pl} & $-$ \texttt{pf} \\
\midrule
All 21 conditions
  & $+15.8\pm1.2$
  & $+1.6\pm0.7$
  & $-3.2\pm0.8$ \\
\midrule
Medium Maze
  & $+4.2\pm1.9$
  & $+2.4\pm1.6$
  & $+0.2\pm1.0$ \\
Diverse Maze
  & $+17.6\pm1.2$
  & $+9.6\pm2.5$
  & $+0.4\pm4.3$ \\
PushObj
  & $+23.9\pm1.7$
  & $-1.9\pm0.9$
  & $-3.9\pm1.5$ \\
PushT
  & $+15.8\pm1.8$
  & $+3.3\pm0.9$
  & $-5.2\pm1.2$ \\
\midrule
Compound (7)
  & $+31.5\pm3.4$
  & $+8.2\pm1.3$
  & $+0.8\pm1.4$ \\
\bottomrule
\end{tabular*}
\end{table}

The relative performance depends strongly on the type of shift. Under
dynamics changes, our method is comparable to \texttt{predfirst}
($+0.5\pm1.3$), whereas the gap is larger for appearance, shape, and
unshifted conditions ($-4.1\pm0.9$). The strongest gains occur under
large dynamics changes. With $50\times$ damping in Medium Maze, our
method reaches 66.0\% success compared with 59.2\% for
\texttt{predfirst} and 46.4\% for Frozen. Under blur combined with
$50\times$ damping, it reaches 65.6\%, compared with 56.0\% and 46.8\%,
respectively. As shown by the ablation in Sec.~\ref{sec:ablation}, these
improvements arise from different residual pathways across environments:
output-side correction dominates in Medium Maze, whereas the action
residual is substantially more important under the PushT controller shift.

The same trend is more pronounced under compound shifts. Across the seven
compound conditions, our method improves substantially over Frozen and
\texttt{predlast}, while remaining comparable to \texttt{predfirst}
($+0.8\pm1.4$). It achieves the highest success rate on four of the seven
conditions and outperforms \texttt{predlast} on all seven. In contrast,
\texttt{predfirst} retains a clearer advantage on several pure appearance
shifts, particularly in PushT, suggesting that direct adaptation of the
pretrained representation remains beneficial when the shift primarily
affects visual encoding. The red-block condition remains challenging for
all methods, with success below 15\%.

Figure~\ref{fig:trajectories} provides qualitative examples under four
compound shifts. These trajectories illustrate the behavior observed in
the aggregate results; in particular, our method successfully reaches the
goal in the illustrated $50\times$ damping episode where the other
adaptation strategies fail.

\subsection{Robot Manipulation: OGBench-Cube}
\label{sec:exp_cube}
\noindent\textbf{Task and world model:}
To test transfer to a 3-D manipulation task and a world model with a
different encoder, we use the single-cube task of OGBench, in which a
UR5e arm with a parallel gripper must move a cube to a target position.
Actions are 5-dimensional (end-effector displacement, yaw, gripper
command), observations are $224\times224$ images from a fixed front
camera, and there is no proprioceptive input. We use the dataset
released with LeWorldModel~\cite{maes2026leworldmodel} and train a
DINO-WM world model on 2{,}000 episodes with the reference recipe of
\texttt{stable-worldmodel}: a frozen DINOv2-small encoder, a 6-block
causal ViT predictor over $K=3$ frames, frames every 5 simulation steps
with 5 actions per block.\newline\newline
\textbf{Evaluation protocol and shifts:}
We follow the LeWorldModel protocol: the start state is a random state
of a dataset trajectory, the goal image is the frame 25 steps later, and
the two preceding frames and executed action blocks form the initial
context. An episode succeeds if the cube comes within 4\,cm of the goal
position. Planning uses the CEM settings of
Sec.~\ref{sec:exp_adajepa} with DINO-WM's cost (mean squared distance
between predicted and goal patch tokens), for up to 10 cycles. Three
shifts, absent from training data, are applied through the simulator: a
\emph{lighting shift} (intensity 0.7 to 0.3), a \emph{viewpoint shift}
($10^\circ$ camera rotation in yaw and pitch), and a \emph{dynamics
shift} (arm joint position gain halved).\newline\newline
\textbf{Baselines:}
The four arms are defined as in Sec.~\ref{sec:exp_adajepa}. Because the
DINOv2~\cite{oquab2024dinov2} encoder is frozen and has no projection
head, \texttt{predlast} updates the last predictor block and its final
LayerNorm and \texttt{predfirst} the first predictor block (3.23M
parameters each); ours updates the residual modules without
proprioceptive terms (99{,}310 parameters). Each cell is evaluated on 50
episodes for each of 3 seeds.\newline\newline

\textbf{Results:}
Table~\ref{tab:cube} summarizes performance on the four OGBench-Cube
conditions. Without shift, Frozen achieves 62.7\% success,
\texttt{predfirst} 64.0\%, and both \texttt{predlast} and our method
60.7\%. Averaged across all conditions, our method reaches 60.0\%,
compared with 58.7\% for Frozen, 57.8\% for \texttt{predlast}, and
58.0\% for \texttt{predfirst}, while adapting only 99{,}310 parameters
versus 3.23M for the AdaJEPA variants. Across the twelve
(condition, seed) pairs, our method differs from Frozen by
$+1.3\pm1.5$, from \texttt{predlast} by $+2.2\pm1.2$, and from
\texttt{predfirst} by $+2.0\pm1.8$.

\begin{figure*}[t]
\vspace*{5pt}
\centering
\includegraphics[width=0.75\textwidth]{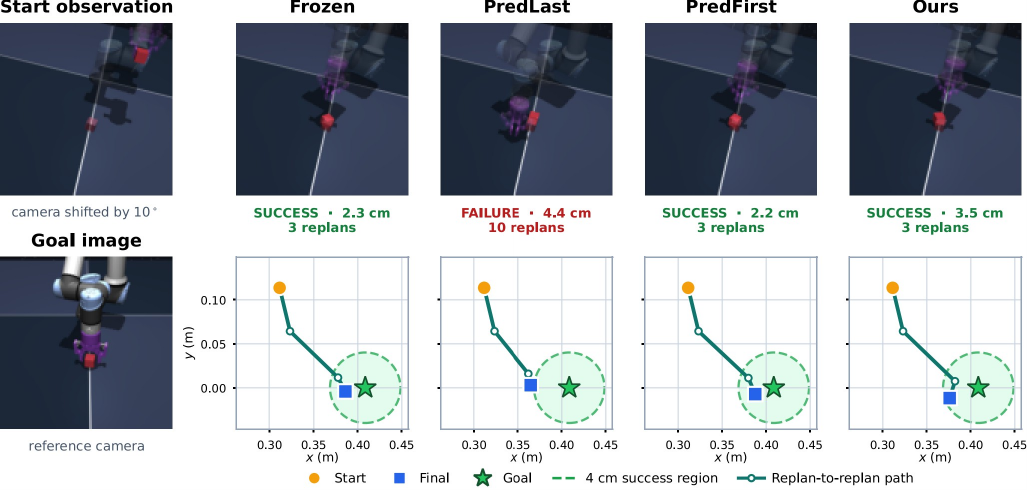}
\caption{Example OGBench-Cube episode for DINO-WM under a
\(10^\circ\) camera shift. All methods use the same start state, goal,
and random seeds. PredLast executes ten replanning steps; however, the
cube undergoes negligible displacement after the first two, causing the
subsequent trajectory markers to overlap near its final position.}
\label{fig:cube_dinowm_camera10_trajectory}
\vspace{-0.5cm}
\end{figure*}

\begin{table}[t]
\centering
\small
\caption{OGBench-Cube with our DINO-WM: success rate (\%), reported as
mean\,$\pm$\,standard deviation over 3 seeds of 50 episodes.
\texttt{pl}: \texttt{predlast}; \texttt{pf}: \texttt{predfirst}.
Bold indicates the best method per condition.}
\label{tab:cube}
\begin{tabular*}{\columnwidth}{
  @{\extracolsep{\fill}}
  lcccc
  @{}
}
\toprule
Condition & Frozen & \texttt{pl} & \texttt{pf} & Ours \\
\midrule
No shift
  & 62.7\pmstd{3.1}
  & 60.7\pmstd{3.1}
  & \textbf{64.0}\pmstd{2.0}
  & 60.7\pmstd{5.0} \\

Light 0.3
  & 58.7\pmstd{4.2}
  & 60.0\pmstd{4.0}
  & 58.7\pmstd{4.6}
  & \textbf{61.3}\pmstd{5.0} \\

Camera $10^\circ$
  & 54.0\pmstd{5.3}
  & 51.3\pmstd{5.0}
  & 52.0\pmstd{5.3}
  & \textbf{56.0}\pmstd{2.0} \\

Arm gain 0.5$\times$
  & 59.3\pmstd{5.8}
  & 59.3\pmstd{5.8}
  & 57.3\pmstd{3.1}
  & \textbf{62.0}\pmstd{2.0} \\
\midrule
\textit{Average}
  & \textit{58.7}
  & \textit{57.8}
  & \textit{58.0}
  & \textbf{60.0} \\
\bottomrule
\end{tabular*}
\end{table}

Under test-time shift, our method is the only adaptation strategy that
improves over Frozen in all three conditions: 61.3\% vs.\ 58.7\% under
the lighting shift, 56.0\% vs.\ 54.0\% under the camera shift, and
62.0\% vs.\ 59.3\% under the arm-gain shift. This corresponds to $+2.4\pm1.4$ relative to
Frozen, whereas both AdaJEPA variants fall below the frozen model under
the viewpoint shift. Figure~\ref{fig:cube_dinowm_camera10_trajectory}
shows a representative episode under this condition and the resulting
replan-to-replan cube trajectory. Overall, these results show that the
same interface-level adaptation principle transfers to a patch-token
world model and a 3-D manipulation setting, although the performance
differences remain modest.

\section{Residual Pathway Ablation}
\label{sec:ablation}

Table~\ref{tab:ablation} analyzes the contribution of each residual pathway under two compound
shifts: Medium Maze with blur $\sigma=2$ and damping $50\times$, and
PushT with blur $\sigma=2$ and doubled controller gain $k_v$. Starting
from the full Sandwich-Residual configuration, we remove either the
action residual $r^{a}$, both output residuals $r^{\mathrm{out}}$, or
both input-side residuals $r^{\mathrm{in}}$. All remaining adaptation
and planning settings are unchanged.

\begin{table}[h]
\centering
\small
\caption{Residual-pathway ablation under compound shifts. Success rates
(\%) are reported as mean\,\(\pm\)\,standard deviation over five seeds
of 50 episodes. Parentheses indicate the change relative to the full
configuration.}
\label{tab:ablation}
\begin{tabular}{lcc}
\toprule
Configuration
& Medium Maze
& PushT \\
\midrule
Full
& \textbf{65.6}\pmstd{6.2}
& \textbf{87.6}\pmstd{3.0} \\

\(-r^{a}\)
& 63.6\pmstd{3.8}\,{\scriptsize(\(-2.0\))}
& 77.6\pmstd{3.0}\,{\scriptsize(\(-10.0\))} \\

\(-r^{\mathrm{out}}\)
& 56.4\pmstd{2.2}\,{\scriptsize(\(-9.2\))}
& 72.0\pmstd{5.1}\,{\scriptsize(\(-15.6\))} \\

\(-r^{\mathrm{in}}\)
& 64.4\pmstd{1.7}\,{\scriptsize(\(-1.2\))}
& 71.2\pmstd{1.8}\,{\scriptsize(\(-16.4\))} \\
\bottomrule
\end{tabular}
\end{table}

The ablation reveals different adaptation mechanisms across the two
environments. In Medium Maze, output-side correction accounts for most
of the improvement: removing $r^{\mathrm{out}}$ reduces success from
65.6\% to 56.4\%, whereas retaining only the output residuals
(\(-r^{\mathrm{in}}\)) achieves 64.4\%, close to the full configuration.
Removing only the action residual has little effect, yielding 63.6\%.
This indicates that the compound shift can largely be compensated for
after prediction rather than through action remapping.

In PushT, the residual pathways are substantially more complementary.
Removing the 210-parameter action residual reduces success from 87.6\%
to 77.6\%, consistent with the controller shift directly modifying the
effect of commanded actions. Removing either all output residuals or all
input-side residuals further reduces success to 72.0\% and 71.2\%,
respectively. Thus, neither side of the predictor alone is sufficient
to recover the full adaptation performance.

Overall, the results show that the dominant correction pathway depends
on the test-time shift. Output-side adaptation is nearly sufficient in
Medium Maze, whereas PushT benefits from jointly correcting both the
predictor inputs and outputs. This supports the full Sandwich-Residual
configuration as a fixed adaptation mechanism when the source of the
test-time mismatch is not known in advance.

\section{DISCUSSION AND CONCLUSION}
\label{sec:conclusion}

\subsection{Limitations}
Sandwich-Residuals assume that the pretrained representation remains sufficient
after test-time shift and that the proprioceptive observation mapping remains
unchanged; we also evaluate shifts that are approximately stationary within an
episode. Our experiments are limited to simulation; real-robot deployment
additionally introduces sensing and actuation noise, latency, and potentially
time-varying dynamics. The residual architecture and learning rates are fixed
across tasks, which may partly explain the remaining gap under visual shifts.
The OGBench-Cube experiments use a reduced training and evaluation setting, and
the observed gains are modest. Finally, reducing the number of trainable
parameters does not proportionally reduce adaptation cost, since gradients for
the input-side residuals must still propagate through the frozen predictor.

\subsection{Discussion}
The results suggest that test-time mismatch does not always require modifying
the internal dynamics model. When the pretrained representation and transition
structure remain useful, adaptation can instead recalibrate predictor inputs or
correct its outputs. The residuals should therefore be viewed as lightweight
compensators rather than explicit system-identification modules: they need only
make the frozen model useful for planning, not recover the true changed dynamics.

The ablation shows that the dominant correction pathway depends on the shift.
Medium Maze is largely corrected after prediction, whereas the PushT controller
shift places greater importance on the action pathway. Appearance shifts are
harder because information distorted or lost inside the frozen encoder cannot be
fully recovered downstream.

\subsection{Conclusion and future directions}
Sandwich-Residuals keep the pretrained world model frozen and learn only small
corrections around its predictor from self-supervised transition error. Across
the 21 primary AdaJEPA conditions, our method achieves \(1.3\times\) the success
rate of the frozen model while retaining 95\% of the strongest AdaJEPA variant's
performance and adapting 97--99\% fewer parameters. Under compound shifts, this
increases to \(1.9\times\) the frozen-model success rate, with performance
comparable to adapting the first predictor block. The same principle transfers
to DINO-WM, where our method is the only adapted variant to improve over the
frozen model under all three tested shifts. Thus, interface-level adaptation can
recover substantial performance without modifying pretrained weights or
selecting a predictor block to update.

Future work could retain residuals across episodes for persistent shifts or
introduce corrections earlier in the visual encoder to better address
representation shifts. In addition, evaluating these extensions on a physical manipulator is
an important next step.

\bibliographystyle{IEEEtran}
\bibliography{references}

\end{document}